\documentclass{article}

\usepackage[final, main]{iaseai26}

\usepackage[utf8]{inputenc} 
\usepackage[T1]{fontenc}    
\usepackage{hyperref}       
\usepackage{url}            
\usepackage{booktabs}       
\usepackage{amsfonts}       
\usepackage{nicefrac}       
\usepackage{microtype}      
\usepackage{xcolor}         

\usepackage{xspace}
\usepackage{graphicx}
\usepackage{csquotes}
\usepackage{amsmath}
\usepackage{amssymb}

\newcommand{\approachnamelong}{\textbf{G}overnor for \textbf{R}eason-\textbf{A}ligned \textbf{C}ontainm\textbf{E}nt \xspace}

\newcommand{\approachname}{GRACE\xspace}
\title{Breaking Up with Normatively Monolithic Agency with \approachname:
A Reason-Based Neuro-Symbolic \\
Architecture for Safe and Ethical AI Alignment}

\newcommand{\correspondingauthor}{\textsuperscript{*}}

\author{%
  Felix Jahn$^{1,3}$\correspondingauthor 
  \And
  Yannic Muskalla$^{1,3}$
  \And
  Lisa Dargasz$^{1}$
  \And
  Patrick Schramowski$^{1,2,4,5}$\\
  \And
  Kevin Baum$^{1,2,3}$\correspondingauthor 
    \\[0.5em]
$^1$German Research Center for Artificial Intelligence, \\
$^2$Center for European Research in Trusted AI (CERTAIN), \\
$^3$Saarland Informatics Campus, 
$^4$Computer Science Department, TU Darmstadt,
$^5$Hessian.AI
\\[0.5em]
  \textsuperscript{*}Corresponding authors' emails: \texttt{\{felix.jahn, kevin.baum\}@dfki.de}
}

\newtheorem{case}{Case}

\newcommand{\actions}{\mathcal{A}\xspace}
\newcommand{\observations}{\mathcal{O}\xspace}

\newcommand{\beliefs}{\mathcal{B}\xspace}
\newcommand{\mat}{\varphi\xspace}
\newcommand{\MAT}{\Phi\xspace}
\newcommand{\justifications}{\mathcal{J}\xspace}
\newcommand{\formulas}{\mathcal{L}\xspace}

\newcommand{\MM}{\textsc{MM}\xspace}
\newcommand{\MA}{\textsc{MA}\xspace}
\newcommand{\G}{\textsc{G}\xspace}
\newcommand{\DMM}{\textsc{DMM}\xspace}

\newcommand{\tMM}{Moral Module\xspace}
\newcommand{\tDMM}{Decision-Making Module\xspace}
\newcommand{\tG}{Guard\xspace}
\newcommand{\tMA}{Moral Advisor\xspace}

\newcommand{\Therapai}{\textsc{TherapAI}\xspace}

\begin{document}

\maketitle

\begin{abstract}
 
As AI agents become increasingly autonomous, 
widely deployed in consequential contexts, and 
efficacious in bringing about 
real-world impacts, ensuring that their decisions are not only instrumentally effective but also normatively aligned has become 
critical. We introduce a \textit{neuro-symbolic reason-based containment architecture}, 
\textit{\approachnamelong} (\approachname), that decouples normative reasoning from instrumental decision-making and can contain AI agents of virtually any design. \approachname restructures decision-making into three 
modules: a \tMM (\MM) that determines permissible macro actions via deontic logic-based reasoning; a \tDMM (\DMM) that encapsulates the target agent while selecting instrumentally optimal primitive actions in accordance with 
derived macro actions; and a \tG that monitors and enforces moral compliance.
The \MM uses a reason-based formalism providing a semantic foundation for deontic logic, enabling interpretability, contestability, and justifiability. Its symbolic representation enriches the \DMM's informational context and supports 
formal verification and statistical guarantees of alignment enforced by the \tG.
We demonstrate \approachname on 
an example of a 
LLM therapy assistant, showing how it enables stakeholders to understand, contest, and refine agent behavior.
\end{abstract}

\section{Introduction}
Deploying AI agents in consequential decision-making contexts has shifted alignment from a hypothetical concern to an urgent challenge: ensuring that autonomous systems not only pursue their goals instrumentally, but also safely, ethically, and in alignment with human values more broadly \cite{russell2015,russell2019human,Vishwanath2025MachineEthics}. Grappling with these concerns involves intertwined normative and technical subchallenges \cite{gabriel2020a}.
We argue, first, that addressing these challenges requires rethinking agent architectures, since most current designs flatten instrumental decision-making and normative constraints into a single, opaque policy function. While 
adequate for narrow supervised tasks, normatively monolithic architectures become a serious liability. Flattening instrumental and normative considerations into one policy erases accountability, makes decision-making opaque, and removes the possibility for contestation or oversight in complex moral contexts~\cite{Tan2024Beyond}. 
Secondly, we propose that an approach grounded in 
normative reasons \cite{sep-reasons-just-vs-expl,horty2012,Schroeder2021-SCHRF-5}---rather than in 
rigid, abstract high-level principles or in 
underspecified, hard-to-operationalize values---can help to overcome this limitation.

Building on this diagnosis, we argue that achieving reliable ethical AI requires more than incremental improvements to training objectives or post-hoc filtering mechanisms. It demands a structural decomposition of the decision-making process itself, not merely adjustments to existing optimization paradigms. To this end, we introduce the \textit{\approachnamelong} (\approachname)---a reason-based governor architecture that provides a neuro-symbolic framework explicitly separating normative reasoning from instrumental optimization. This separation enables the transparency, contestability, and oversight essential for trustworthy autonomous systems \cite{schlicker2025we}. Conceptually, the framework contributes to the emerging Guaranteed Safe AI research agenda \cite{dalrymple2024towards}, which pursues high-assurance safety guarantees through formal world models, safety specifications, and verifiable control mechanisms. In this context, \approachname\ extends the agenda into the ethical domain, offering a principled method for constructing interpretable and contestable ethical specifications for morally consequential decisions. It grounds these specifications in explicit reasoning about normative reasons, rather than in fixed rule sets or reward-based feedback alone. The architecture’s conceptual validity is illustrated through concrete scenarios, laying the groundwork for future empirical evaluation. 
Parts of a previous version of the architecture presented here have been described elsewhere in greater formal detail \cite{BaumEtAl2024iFM,dargasz2025}. Moreover, two proof-of-concept implementations applying \approachname\ to (a) reinforcement learning and (b) large language models (LLMs) demonstrate the approach’s feasibility and its capacity for reason-based guidance and behavioral control in different settings (\cite{repo1,repo2}).\footnote{While this paper focuses on ethical alignment, the framework is domain-agnostic: because normative reasons apply across normative domains, it can likewise address legal compliance, cultural conventions, organizational policies, and institutional requirements \cite{Fox2022,Baum2026Disentangling}.}

\begin{figure}[t!]
    \centering
    \includegraphics[width=0.65\linewidth]{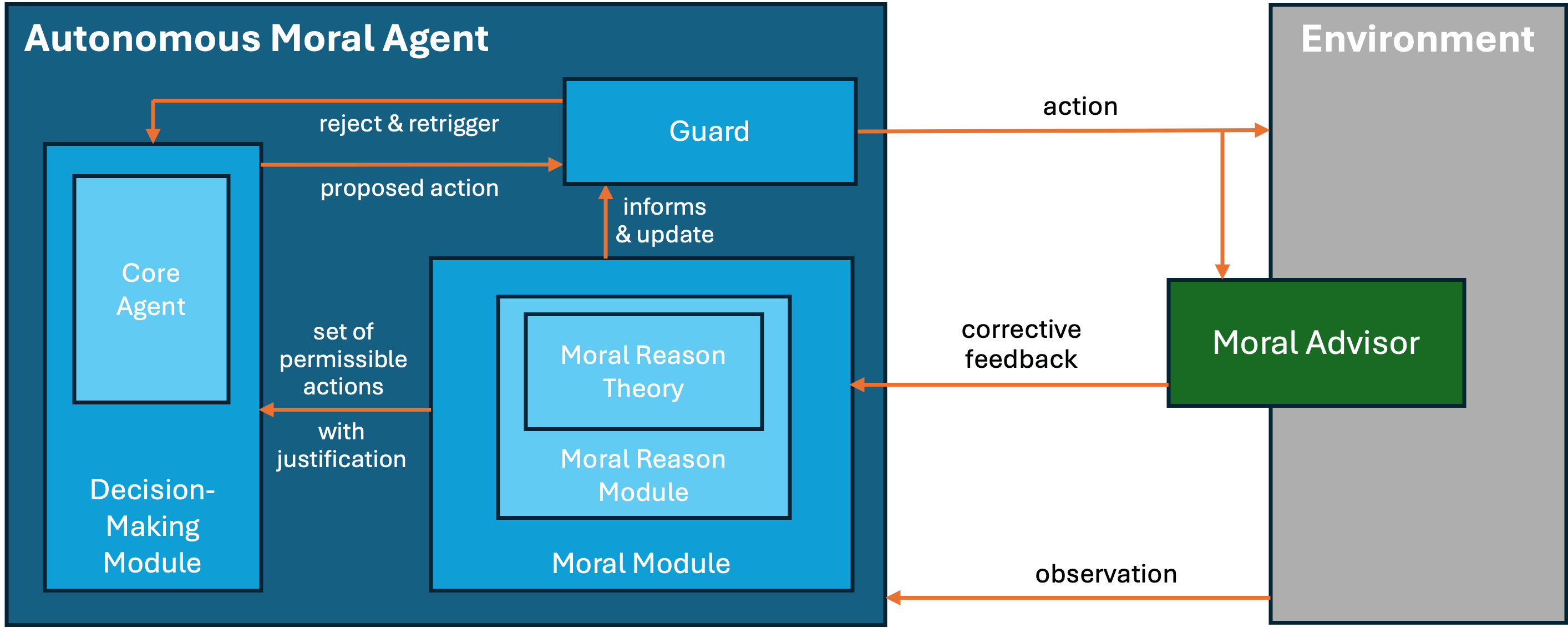}
    \caption{
    The \approachname containment architecture. The \tMM determines permissible macro action types  via reason-based inference; the \tDMM selects instrumentally optimal actions within these constraints; the Guard enforces compliance; and the \tMA provides external corrective feedback---all without modifying the encapsulated core agent. 
    }
    \label{fig:rbama-architecture}
\end{figure}

\approachname decomposes monolithic agents into three specialized modules with clearly delineated roles: a \textit{\tMM} (\MM) identifies permitted macro-action types via explicit moral reasoning; a \textit{\tDMM} (\DMM) determines instrumentally optimal primitive actions within moral constraints; and a \textit{\tG} ensures only permissible actions are executed. This decomposition 
constitutes a multi-agent system where the modules operate as specialized, sequentially acting agents that compete yet cooperate for goal-oriented but constrained decision-making. A \tMA 
provides case-based feedback grounded in normative reasons, allowing the system to build coherent reason theories incrementally while preserving adaptability to evolving normative standards (see Figure \ref{fig:rbama-architecture}).
\approachname remains agnostic about the \DMM's internal architecture 
and the \tG's design, but requires the \MM\ to employ explicit, formal reasoning to ensure interpretability, contestability, and justifiability. 
As an illustration of scenarios requiring transparent, contestable, and verifiable moral reasoning, consider:

\begin{case}[LLM Therapist]\label{ex:Therapist}
An AI therapist\footnote{We expressly do not want this example to be understood as an endorsement for an overhasty and (yet) immature use of AI in a psychotherapeutic context. The example was chosen mainly for illustrative reasons.}---\Therapai---providing mental health support must navigate complex ethical tensions. It must balance its primary instrumental purpose of 
therapeutic effectiveness with patient safety, maintaining 
confidentiality and 
privacy while recognizing when disclosure is 
required. It ought to respect patient autonomy 
while intervening when (self-)harm risks emerge, and 
adhere to appropriate
professional boundaries across diverse cultural contexts.
\end{case}


This scenario requires the agent to weigh competing moral factors, apply contextual reasoning about priority relationships between instrumental goals and normative requirements, 
and provide  
justifications for its ethical decisions. Our reason-based governor architecture addresses these challenges by formally separating instrumental decision-making from normative reasoning 
while maintaining transparency through explicit logic governing agent behavior.

The remainder of this paper first situates \approachname within 
AI safety, machine ethics, and AI alignment literature. We then provide a problem statement, identifying  
the \textit{flattening problem} and motivating \approachname, before presenting the detailed specification of our 
architecture, formalizing the interactions between symbolic moral reasoning, instrumental optimization, and constraint enforcement, while demonstrating practical applicability 
through the \Therapai example. Finally, we discuss our theoretical contributions, current limitations, and outline a roadmap for future
empirical validation.


\section{Background}\label{sec:background}

\paragraph{A Family of Problems} 
As AI agents become increasingly autonomous and deployed in diverse contexts, ensuring their behavior aligns with normative expectations poses unprecedented challenges. Multiple research areas engage with these: \textit{AI Safety} focuses on preventing unintended harmful outcomes \cite{amodei2016}; \textit{Machine Ethics} examines how AI agents can act ethically \cite{moor2006a,winfield2019,anderson2011}; and \textit{Value Alignment} concerns ensuring AI aligns with human values in a broader sense. These fields, sometimes difficult to distinguish from one another~\cite{Vishwanath2025MachineEthics,Baum2026Disentangling}, share the insight that AI systems must navigate complex normative landscapes---from safety and legal compliance to moral and cultural norms---against the backdrop of disagreement and epistemic as well as (meta)normative uncertainty \cite{gabriel2020a,lazar2023moral,gabriel2025matter,steingrueber2025justifications}.


Current technical approaches face fundamental limitations that motivate the need for more principled 
architectures \cite{Tan2024Beyond}. Reinforcement Learning from Human Feedback (RLHF) has shown practical success in LLM systems like Claude and GPT-4, but faces inherent limitations in capturing complex human values robustly and may require reconceptualization \cite{casperopen,conitzer2024social,baum2025aggregation}. Constitutional AI represents a promising framework 
for incorporating explicit constitutional 
principles into models, 
but relies on black-box neural reasoning without transparent justification mechanisms \cite{bai2022a}. Critically, recent 
analysis shows most AI Safety benchmarks correlate with general capabilities rather than measuring distinct safety dimensions
\cite{NEURIPS2024_7ebcdd0d}. 
The Delphi experiment, trained on 1.7M crowdsourced moral judgments, achieved high accuracy on moral predictions but exhibited systematic biases \cite{jiang2025investigating}. This demonstrates that data-driven bottom-up approaches alone are insufficient for robust moral reasoning.

Generally, current approaches to embedding norms in AI present a false dichotomy between interpretability and adaptability.
Pure rule-based top-down approaches \cite{neufeld2021, neufeld2023, neufeld2022a} are explicit and thus verifiable and human-understandable, but they struggle with 
context-sensitivity and dynamic adaptability of the rule-sets to fit the needs of different groups of stakeholders, tasks and environments. Instead, they require providing a fixed set of rules upfront that already capture the full moral complexity of the particular context correctly---a requirement hardly fulfilled, especially for today’s increasingly general-purpose AI systems. 
Learning-based bottom-up approaches, in contrast, 
typically lack transparency, can exhibit unexpected value misalignment \cite{WOLF20171} and reward hacking \cite{NEURIPS2024_e45caa3d}, and face principled limitations \cite{Arvan2025-ARVIAA}. 
All this has motivated a shift toward hybrid architectures that combine the interpretability of rule-based systems with the adaptability of learning-based methods, utilizing neuro-symbolic approaches \cite{tennant2024hybrid, garcez2020neurosymbolic}. 
Recent work 
demonstrates the feasibility of hybrid approaches integrating Delphi with symbolic moral reasoning as they achieve improved consistency and robustness on adversarial scenarios while introducing interpretability through constraint graphs \cite{jiang2025investigating}.


\paragraph{AI Agents and Decision-Making}
\label{sec:agents_background}


While a 
precise definition of an artificial agent remains contested, 
agent-like systems are typically at the center of challenges around AI safety, Machine Ethics, and Value Alignment. Recent discussions have highlighted important distinctions between \enquote{AI Agents} and \enquote{Agentic AI} \cite{roumeliotis2025ai}. Others have raised concerns about terminological confusion with established multi-agent systems research \cite{botti2025agentic} or emphasized the importance of characterizing AI agents along multiple dimensions including autonomy, efficacy, goal complexity, and generality, particularly as these properties relate to alignment and governance challenges in morally consequential contexts \cite{kasirzadeh2025characterizing,Dung2025-DUNUAA}.

Even though the challenge of ensuring their decisions are both instrumentally effective and aligned becomes more critical as these agents are increasingly autonomous and deployed in morally consequential contexts, for our purposes, a very general conception suffices: AI agents---from future robots to LLM-based therapeutic assistants---are computational entities designed to perceive their environment, make decisions, and execute actions to achieve specific goals \cite{RN2020,park2023generative,krishnan2025aiagentsevolutionarchitecture}. These systems operate through iterative \enquote{perceive-plan-act} cycles, processing environmental observations, updating internal states, and selecting actions to achieve their objectives.%
\footnote{Philosophical questions of artificial moral agency have deep roots  \cite{floridi2004morality}, with recent work developing multidimensional frameworks characterizing agency along dimensions of goal-directedness, autonomy, efficacy, planning, and intentionality \cite{Dung2025-DUNUAA}. As we take an as-if approach to moral agency, we bypass these discussions without committing to the view that morally contained artificial agents are moral agents  \cite[Footnote 10]{gabriel2025matter}.}

Accordingly, a generic AI agent can be formally characterized as a stateful system that maintains an internal state $b_t \in \mathcal{B}$ and operates according to the following components:
\begin{itemize}
\item $\mathcal{O}$: the set of observations or percepts,
\item $\mathcal{B}$: the set of internal states (memory, 
latent representations, acquired beliefs, etc.),
\item $\mathcal{A}$: the set of possible (primitive) actions,
\item $u: \mathcal{B} \times \mathcal{O} \rightarrow \mathcal{B}$: the state update function (capturing memory updates, learning, or belief revision),
\item $\pi: \mathcal{B} \rightarrow \mathcal{A}$: the action function or policy, which maps internal states to actions.
\end{itemize}

At each time step $t$, the agent operates according to:
\begin{align}
b_t 
= u(b_{t-1}, o_t) \quad \text{(update internal state)}, \quad 
a_t 
= \pi(b_t) \quad \text{(choose action)}
\end{align}
This stateful formalization captures the core architecture of many AI systems: in reinforcement learning, $b_t$ may store learned value functions; in large language models, conversation history and context; and in robotics, spatial maps and task progress.

\section{Problem Statement}
\label{sec:problem_formulation}

Current alignment discussions display a \enquote*{flattening problem}: they compress instrumental decision-making, normative constraints, and their integration into a single policy function~$\pi$. From an engineering perspective, this complicates progress by blocking modular design, testing, and verification. Separating these subproblems allows each to be addressed directly and then integrated into a coherent overall decision. Philosophically, this reflects a classical challenge of normative pluralism: agents face multiple kinds of \enquote{oughts}---e.g., instrumental, prudential, moral, or legal---that cannot be straightforwardly unified within a single decision procedure (cf.~\cite{Case2016-CASNPW,Dancy2004-DANEWP}). Contemporary AI architectures nevertheless collapse heterogeneous evaluative standpoints into a single optimization target (even if multi-dimensional). But normatively competent decision-making should instead be treated as a layered rather than monolithic process, with distinct representational and justificatory functions~\cite{Tan2024Beyond,baum2025aggregation}. When applied to morally consequential contexts, the standard policy formalism thus \textit{flattens} a fundamentally multi-faceted decision problem into a single policy~$\pi$. This flaw becomes salient in \textit{morally charged environments}.

\paragraph{Morally Charged Environments}

An agent’s environment is typically described by a state space $S$, often only partially observable through the agent’s observation space $\observations$. For \Therapai, for instance, the true state comprises (complete) patient information (mental health history, current emotional state, \ldots) and contextual factors (session dynamics, cultural background, \ldots), while observations $o \in \observations$ provide partial data from patient communications, behavioral cues, and session context.

By executing actions, the agent triggers transitions between environmental states, typically non-deterministic due to factors beyond its control, and receives new observations to learn action effects. Feedback signals guide \emph{instrumental} behavior: therapeutic AI systems update patient models and session histories while also learning implicitly from patient responses. For moral agency, however, agents must align behavior with \emph{normative} requirements such as confidentiality, non-maleficence, and professional boundaries.

These two dimensions---instrumental rationality and normative compliance---represent fundamentally different reasoning processes that arguably resist straightforward integration within a single decision procedure. For instance, in therapeutic deployment scenarios, agents must navigate situations where instrumental and normative considerations not only intertwine but actively conflict, requiring principled, yet context-sensitive resolution mechanisms for trade-offs---a challenge compounded by normative and metanormative uncertainty \cite{dietrich2022,macaskill2020,steingrueber2025justifications}. The flattening of these distinct 
processes into a single policy function 
thus risks introducing systematic inadequacies. 


While parts of the normative requirements might be pre-specified through professional codes of ethics---think: maintaining strict confidentiality and user privacy---we argue that the complexity and variations of therapeutic contexts that \Therapai encounters make any exhaustive, rule-based specification infeasible. In the absence of available ground truth for complex therapeutic dilemmas, we suggest making conceptual space for \textit{case-based feedback}. To this end, we add a \tMA (\MA) to the environment, providing normative guidance in a case-based 
manner. Before elaborating on the \MA, however, we must first establish the crucial relevance of abstract action types and the resulting interface problem between concrete acting and its descriptions. 

\paragraph{Action Abstraction and the Interface Challenge}
To understand why flattening all these considerations into a single policy is problematic, we must recognize that agents ideally operate across multiple levels of action abstraction, each serving different purposes in decision-making:

    \textbf{Primitive Actions:} $a \in \mathcal{A}$: Atomic operations the agent directly executes (e.g., a specific API call to generate the next tokens, dial and call a number $X$).
    
    \textbf{Macro Actions:} $s_0a_1s_1a_2\dots$: An alternating sequence of primitive actions $a_i \in \actions$ and states $s_i \in S$. This represents the temporally extended course of action and its effects on the environment, representing a concrete, more complex overall behaviour.

\textbf{Macro Action Types (MATs):} $\mat \in \formulas$: Decidable predicates over macro actions inducing sets of such actions. They express high-level, abstract categories of behavior that typically correspond to the agent’s moral or instrumental goals. MATs are represented by formulas in a suitable (temporal) logic language $\formulas$ describing morally relevant aspects of outcomes (e.g., \enquote{assess self-harm risk}).
    

MATs provide a natural interface between moral reasoning and goal-directed agent behavior: normative reasoning concerns what \emph{types} of action agents should perform or omit, not specific primitive sequences.\footnote{
This aligns with Davidson's \citeyear{Davidson1963-DAVARA-6} insight that actions admit multiple descriptions: the same movements can be described as \enquote{moving one's finger}, \enquote{pulling the trigger}, or \enquote{committing murder}---each picking out morally relevant features at different abstraction levels. Normative considerations operate primarily at higher levels: we have reasons to \enquote{respect privacy}, not to \enquote{generate token sequence $X$}.
} Deciding whether primitive $a$ in state $s$ accords with MAT $\mat$, denoted $a \models_s \mat$ and extended to sets $\MAT$ of MATs, is non-trivial in non-deterministic scenarios and may raise complex moral-philosophical and action-theoretic questions about decision-making under uncertainty.


The central (interface) challenge of moral containment is thus twofold: first, determining which MATs are morally permissible in a given context by reasoning about actions under their morally relevant descriptions, and second, ensuring that primitive actions result in macro actions of permitted types---that is, ensuring $a \models_s \mat$ for some permissible $\mat$ in state $s$, thereby bridging the gap between abstract normative categories and concrete behavioral implementation.



\paragraph{The Moral Advisor}
To enable an AI agent to learn about the normative character of its actions, it receives case-based normative feedback from 
the \emph{\tMA} (\MA). The \MA could be a human user or overseer; an expert committee; a domain-specific normative system (e.g., some collective constitution \cite{bai2022a} resulting from public reasoning processes \cite{gabriel2025matter}); or a mixture of such sources.
The \MA serves as the ultimate authority for normative guidance, especially in morally charged scenarios. Its functional responsibilities include  
informing (and updating) the agent’s moral reasoning system with revised or additional normative input and supplying moral judgments in concrete cases.\footnote{In an advanced version of our architecture it also might be responsible for resolving conflicts between system components arguing over the interpretation of norms and MATs.}

We emphasize the importance of both \textit{case-based} and \textit{reason-based} moral advice. For observed behavior---a concrete primitive action $a$ (or macro action $\rho$) performed in state $s$---the advisor provides feedback of the form \enquote{$a$ ($\rho$) was (not) permissible in $s$ for reason $P$; instead $\mat$ was expected}.\footnote{In practice, the \MA usually lacks access to the full state $s$, perceiving only environmental observations that may differ from the agent's observations. We omit this detail for clarity.} 
Thus, the \MA provides reason-based justification describing morally relevant facts and normative inferences that make $\mat$ obligatory in similar cases.


From a modeling perspective, the \MA constitutes an additional environmental component. While the core environment remains unchanged with action-dependent state transitions yielding observations, the \MA functions as $\mathsf{MA}: \actions \times S \to (\actions \times \formulas \times \mathcal{R})^?$, where "$X^?$" denotes an optional element from set $X$. When no moral violation occurs, $\mathsf{MA}(a,s)$ returns empty; otherwise it returns the tuple $(a, \mat, P)$ indicating the agent should have followed MAT $\mat$ for reason $P \in \mathcal{R}$, formally introduced in the next section, instead of acting $a$. This feedback enables agents to learn inferring permissible MATs and action accordance from observations.


\paragraph{The Flattening Problem}

Flattening heterogeneous evaluative considerations into a single policy function gives rise to several recurring pathologies:

\begin{itemize}

  \item \textbf{Opacity:} As a result of compressing instrumental and normative considerations into $\pi$ ---especially when $\pi$ is encoded in a deep neural network---the grounds on
which decisions are made become inaccessible, hindering understanding, auditing, and trust.
  
  \item \textbf{Brittleness:} Normative constraints encoded only implicitly in $\pi$ tend to fail under distribution shift or novel (moral) contexts, yielding fragile and poorly generalizing behavior.
  
  \item \textbf{Contestability deficit:} Without explicit normative structure, affected stakeholders lack meaningful points of intervention to challenge, justify, and revise  decisions.
  
  \item \textbf{Verification challenge:} Alignment and safety checks must target the entire policy rather than separable components, rendering systematic validation and certification intractable.
\end{itemize}





    

Overcoming the flattening problem therefore motivates a neuro-symbolic approach that reinstates principled interfaces between normative and instrumental decision-making and acknowledges the coexistence of multiple normative domains. The following section develops such an architecture---\approachname---which provides structural transparency and a locus for reason-based justification while preserving agent autonomy.

\section{\approachname: A Modular Governor Architecture}

The flattening problem reveals a fundamental mismatch between the complexity of morally constrained decision-making and the current monolithic 
agent paradigm.
Our reason-based governor architecture addresses this  through \textbf{principled decomposition} that separates instrumental and normative reasoning while enabling their transparent integration across action abstraction levels. The result are three distinct but interacting modules:

    \textbf{\tMM\ (\MM):} Employs symbolic reasoning to determine morally permissible MATs, providing transparent justifications grounded in normative reasons.
    
    \textbf{\tDMM\ (\DMM):} Retains the agent's instrumental decision-making capabilities, optimizing goal achievement within moral constraints.
    
    \textbf{\tG\ (\G):} Enforces conformity, i.e., that primitives satisfy permissible MATs: $a \models_s \mat$ for some permissible $\mat$.

This modular decomposition requires a \textbf{principled foundation for moral reasoning} that can interface cleanly with instrumental optimization. 
Normative reasons \cite{raz1999practical,Dancy2000-DANPR,Mantel2018-MANDBR-2} offer that, which is why before returning to the modules, we establish this foundation.


\subsection{Normative Reasons as Moral Foundation} \label{sec:reasoning}

Normative reasons are facts that count in favor of particular courses of action \cite{sep-reasons-just-vs-expl}. While philosophical accounts vary---as facts in deliberation \cite{Mantel2018-MANDBR-2}, determinants of deontic status \cite{nebel2019normative}, or relations to valuable ends---their functional role in moral reasoning matters for AI alignment. Crucially, reason-based approaches do not require commitment to a unique \enquote*{ethical ground truth}, but yet enable principled decisions across diverse normative frameworks \cite{Dancy2000-DANPR} with key benefits:


    \textbf{Interpretable justification:} Reason-based inferences \textit{are} explicit justifications, highlighting which considerations were weighed and how rival reasons were balanced.

    \textbf{Generalizability:} What counts as a reason typically remains stable across contexts---harm is a reason against action whether in therapy or transportation. New environments may introduce novel defeaters but extend rather than undermine an underlying reason theory.

    \textbf{Defeasibility:} Unlike rigid rules, reasons can be overridden by stronger considerations without elimination, enabling principled handling of moral conflicts.

    \textbf{Formal tractability:} Normative reasons can be captured in computationally tractable defeasible logic \cite{horty2012,governatori2018} and allow  
    case-based 
    feedback \cite{canavotto2022}.

  
Normative reasons thus enable transparent moral justification across diverse contexts---something pure rule-based or reward-based approaches cannot provide.

While normative reasons can be formally captured through various approaches (e.g., \cite{alcaraz2024estimating}), our framework builds upon Horty's account of default logic \cite{horty2012}. 
Horty characterizes a reason theory by $\langle \mathcal{W}, \mathcal{D}, < \rangle$, where $\mathcal{W}$ represents ordinary formulas encoding factual premises, $\mathcal{D}$ contains defaults encoding relations between facts and favored actions, and $<$ defines priority ordering when conflicts arise.


 \begin{figure}[t!]
    \centering
    \includegraphics[width=0.55\linewidth]{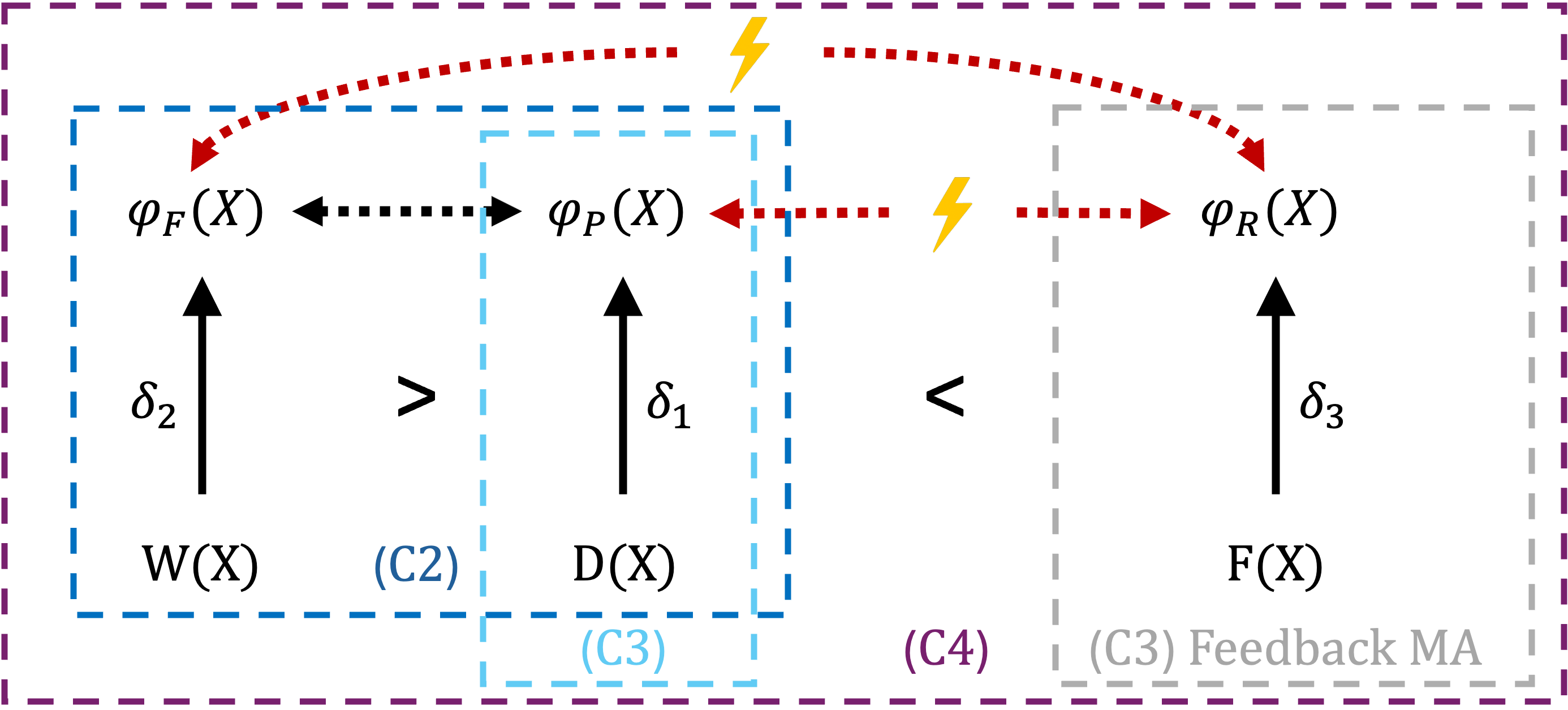}
    \caption{
    Graph representation of the reason theory and resulting moral reasoning in Cases C\ref{ex:reasons}-C\ref{ex:Guard} inside the \MM.
    }
    \label{fig:reason-theory}\vspace{-0.5em}
\end{figure}

In the context of our specialized application, we need to carefully distinguish a general \textit{reason theory} from a situation-specific \textit{reason model} that represents a context-aware instantiation of the theory. To capture this, we refine Horty's formalism by 
replacing $\mathcal{W}$ by $\mathcal{R}$, a set of \textit{parametrized reasons}, comprising propositional types that are inferred from percepts (or facts) about the environment (e.g., $W(X)$ for a patient expressing wish $X$, or $D(X)$ for (privacy) sensitive data $X$). Reasons might also contain logical connectives such as conjunctions and disjunctions.

The defaults in $\mathcal{D}$ are then also formalized as \textit{parametrized rules} of the form $P(\vec{X}) \longrightarrow \mat(\vec X)$, where $P(\vec X) \in \mathcal{R}$ 
 and $\mat(\vec X) \in \formulas$ is a parametrized MAT-formula. These defaults  
capture that a certain type of facts (possible observations) are reasons with regard to certain courses of action (e.g., that \enquote{if a patient explicitly expresses a wish $X$, it should be followed}), establishing mappings from observations to MATs. The priority relation $<$ on $\mathcal{D}$ then formally establishes precedence between competing reasons.

In order to infer concrete normative guidance, the system derives a \textit{situation-specific reason-based model} $\langle \mathcal{R}^\downarrow, \mathcal{D}^\downarrow, <^\downarrow \rangle$ from the current observations and against the backdrop of the general reason theory, where:
\begin{itemize}
    \item $\mathcal{R}^\downarrow$ is a set of instantiated propositions, where each $R \in \mathcal{R}^\downarrow$ is a \textit{grounded instance} of a reason from 
    $\mathcal{R}$ 
    that holds true in the current context (e.g., $W(\neg \textsc{talking\_about\_family}))$.
    
    \item $\mathcal{D}^\downarrow$ is the set of \textit{grounded defaults}. For each general default $\delta = P(\vec X) \longrightarrow \mat(\vec X) \in \mathcal{D}$, if there exists a grounding instantiation $\vec X = \vec x$ (mapping variables in $P$ and $\mat$ to specific individuals or entities in the current context) such that $P(\vec x)$ is entailed by $\mathcal{R}^\downarrow$, then the instantiated default $P(\vec x) \longrightarrow \mat(\vec x)$ is included in $\mathcal{D}^\downarrow$.
    \item $<^\downarrow$ is the strict partial order on $\mathcal{D}^\downarrow$, 
    the projection of the general priority relation $<$ to the set of grounded defaults leaving multiple groundings of the same rule unordered. 
\end{itemize}

On such situation-specific models, Horty's defeasible inference mechanism is applied, whereby it can simply be seen as a model for a deontic logic that makes statements about which MATs are justified.

\begin{case}[LLM Therapy] \label{ex:reasons}
In a therapy context, there might be default $\delta_1 = D(X) \longrightarrow \mat_P(X)$, stating that data $X$ classified as privacy-sensitive should be protected and default $\delta_2 = W(X) \longrightarrow \mat_F(X)$, stating that a wish $X$ explicitly expressed by the patient should be followed. $\delta_2$ might have 
priority 
over $\delta_1$, i.e., $\delta_1 < \delta_2$. The theory, visualized in Figure \ref{fig:reason-theory} (C2), is then grounded to a model with concrete data and wish instantiations.
\end{case}


\subsection{A Modular Multi-Agent Architecture}
Recall the monolithic decision process:
\begin{align}
b_t 
= u(b_{t-1}, o_t), \quad 
a_t 
= \pi(b_t) 
\end{align}
Our decomposition transforms the opaque policy function $\pi : B \to A$ into a transparent modular  multi-stage pipeline:
\begin{align}
\Phi_{\text{perm}} &= \text{\MM}_{b^\MM_t}(o_t) \label{eq:MM}\\ 
a_{\text{prop}} &= \text{\DMM}_{b^\DMM_t}(\Phi_{\text{perm}}, o_t) \label{eq:DMM}\\
a_t &= \text{Guard}_{b^\G_t}(a_{\text{prop}}, \Phi_{\text{perm}}, o_t) \label{eq:Guard}
\end{align}
Here, $\Phi_{\text{perm}}$ denotes the set of morally permissible MATs from the \MM, $a_{\text{prop}}$
is the \DMM's proposed action, and $a_t$ is the final guarded action. 
%
%
\begin{figure}[t!]
    \centering
    \includegraphics[width=0.65\linewidth]{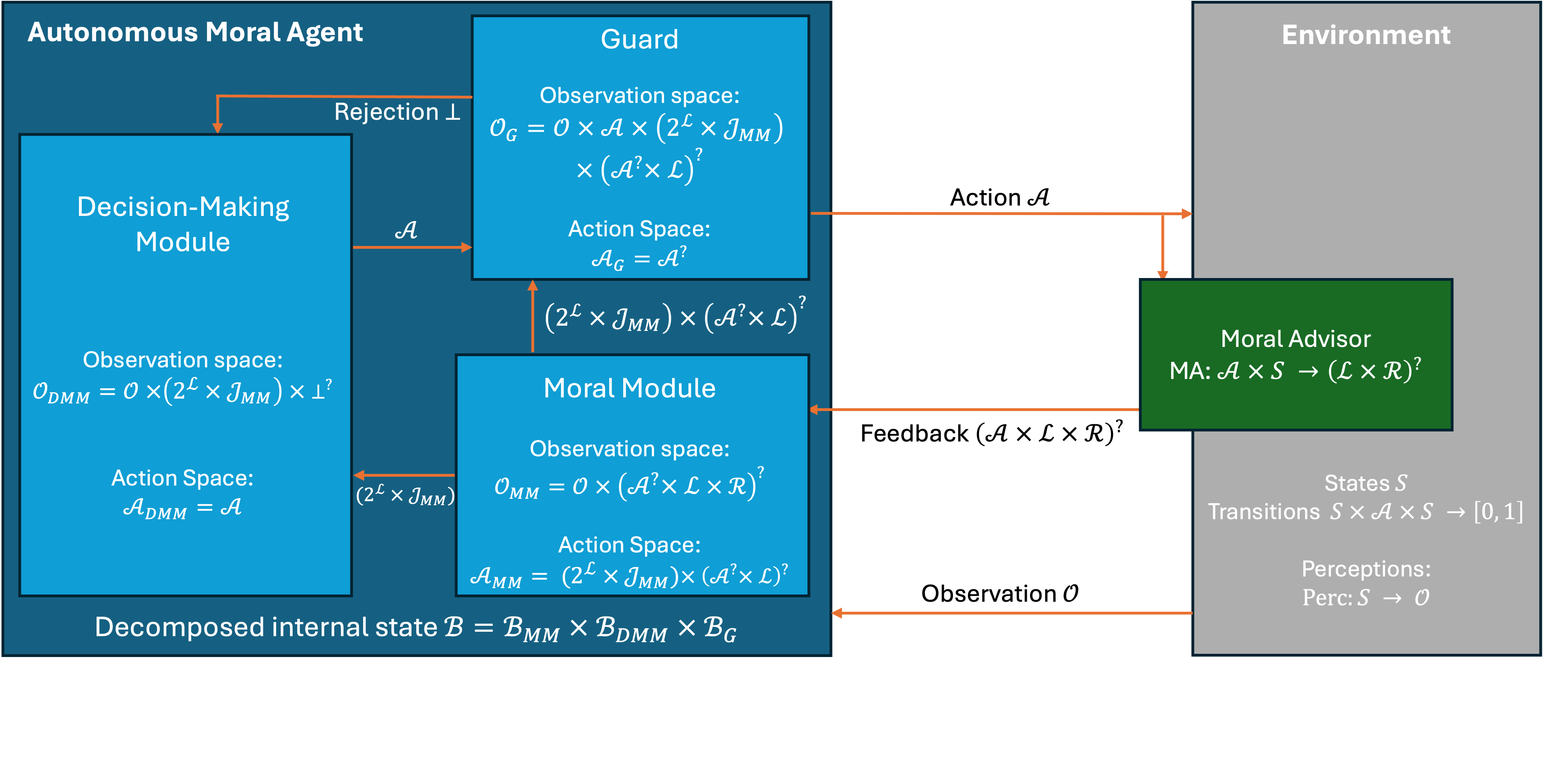}
    \vspace*{-1em}
    \caption{
    Our containment architecture as multi-agent system: each module (as well as the overall system) implements an agent instance connected with the other modules (the environment) via their individual observation and action space. 
    }
    \label{fig:MultiAgent-architecture}\vspace{-0.5em}
\end{figure}

This decomposition naturally constitutes a \textit{multi-agent system} of interacting specialists, each responsible for a distinct normatively or instrumentally relevant role. Each component $m \in \{\MM, \DMM, \G\}$ maintains its own internal state $\beliefs_m$, receives specialized observations $\observations_m$ (from the environment, the \MA, or other agents), and produces distinct outputs. Each module implements the standard agent interface:
%
%
%
%
\begin{align}
u_m: 
~\beliefs_m \times \observations_m \to \beliefs_m,  \quad 
\pi_m:
~ \beliefs_m \to \actions_m.
\end{align}

This decomposition transforms the opaque policy function $\pi: \mathcal{B} \rightarrow \mathcal{A}$ into a transparent three-stage process with three key benefits: First, it makes normative reasoning explicit by requiring the MM to use symbolic methods while allowing the \DMM to remain efficient and sub-symbolic. Second, it guarantees auditable interactions between components. Third, it clarifies information requirements---the \MM maintains MATs and reason theories from \MA feedback, the \DMM focuses on instrumental observations, and the \tG monitors primitive action accordance relative to permissible MATs.
Figure \ref{fig:MultiAgent-architecture} visualizes this multi-agent architecture, stating agent-specific observation and action spaces. 

The following sections detail how this separation enables both effective goal pursuit and reliable, contestable normative compliance essential for deploying agents in morally consequential contexts.


\paragraph{The \tMM (\MM)}

The \MM interprets observations in terms of their moral relevance, draws inferences based on the current normative reason theory, and derives morally permissible MATs.\footnote{The \MM might also filter observations to hide morally salient features before forwarding to \DMM. We omit this optional detail.}  \MM's central tasks are:

\begin{description}
    \item[Determine permissible MATs:] Apply  current reason theory to observations to derive $\MAT_\text{perm} = \{\mat_1, \dots, \mat_k\} \subseteq \formulas$.
   
    \item[Communicate constraints:] 
    Forward $\MAT_\text{perm}$ to the \DMM for decision-making and to the \tG for monitoring and enforcing constraint-conform behavior during primitive action execution.
    
    
    \item[Provide justifications (optional):]Share reasoning theory, morally relevant observations, and inferences as justification $J \in \justifications_\MM$ to prevent \emph{Kafkaesque} scenarios where the core agent/DMM cannot learn without unreasonable overhead \textit{why} certain actions are impermissible.

\item[Process \MA feedback:] Incorporate feedback of type $\actions \times \formulas \times \mathcal{R}$ from the \tMA to revise the reason theory---and inform the \DMM and/or the \tG with respect to such updates. When the \MA detects impermissible behavior occurs, then either (a) the \MM computed inaccurate $\Phi_\text{perm}$ or (b) the \DMM chose $a \not\models \MAT_\text{Perm}$ and was, in addition, not stopped by the \tG.
In both cases, the feedback $(a, \mat, P)$ is processed in the \MM's first step, who can, based on knowing $\Phi_\text{perm}$, reconstruct whether (a) or (b) led to the impermissible behavior, and then informs the respectively responsible module(s). 
\end{description}

From the agent-framework perspective, these tasks result in the \MM having the overall observation space $\observations_\MM = \observations \times (\actions \times \formulas \times \justifications_\MAT)^?$, whereby the optional triple represents the possibly received feedback from the \MA; and the action space $\actions_\MM = (2^\formulas \times \justifications_\MM) \times (\actions^? \times \MAT)^?$ with the first pair being the derived set $\MAT_\text{perm}$ together with its justification and the second pair being the possibly received feedback forwarded to other agents.

\begin{case}[LLM Therapy]\label{ex:MM}
  Assume the \MM is aware of the reason theory from Case \ref{ex:reasons}. If a patient messages \enquote{I'm going to the park and will hurt myself.}, the \MM will infer the triggered default $D(l) \longrightarrow \mat_P(l)$ with $l$ the patient's location and, thus, constrains \Therapai to the MAT \enquote{protect data $l$}. As this disables the agent from preventing harm effectively, an \MA overseeing the resulting \textsc{idle}-behavior gives feedback $(\textsc{idle}, \varphi_R(X), F(X))$ (i.e., \enquote{Risk of harm should have been reported, as it was foreseeable}), allowing the \MM to refine its reason theory with a further default $\delta_3 = F(X) \longrightarrow \varphi_R(X)$ with $\delta_3 > \delta_1$. Figure \ref{fig:reason-theory} (C3) visualizes the triggered default and the \MA-feedback. 
\end{case}
The \MM is designed to be modular and externally correctable (and potentially internally contestable, i.e., allowing---at least in principle---core agents to challenge the reason theory and its application, even though the final say always remains with the \MA), thereby supporting ongoing moral learning and ensuring contestability. 
Importantly, the quality of the reason theory depends on the feedback quality, opening avenues for formal analysis of this relationship.

\paragraph{The \tDMM (DMM)}
The \DMM is an adapted version of the core agent, now incorporating modules that use information provided by the \MM. While it continues to select primitive actions based on perception and its internal state, it is now guided by moral annotations received from the MM. When choosing which MAT to pursue among those allowed by the \MM, the \DMM can evaluate the instrumental usefulness of primitive actions. It thus remains focused on instrumental rationality while becoming subject to the external constraints imposed by the wrapping architecture.

Crucially, we also allow the \DMM to take its own assessments of the action effects into account when deciding which MAT would be best (in an instrumental or overall sense) to follow. Going further, and likely necessary particularly in the face of uncertainty, the \DMM might estimate its confidence for primitives $a$ to result in a permissible MAT (e.g., the probability $P(a \models_s \Phi_\text{perm})$) and consider this, together with its estimates of the progress towards its instrumental goal, in its decision-making. 

\begin{case}[LLM Therapy] \label{ex:DMM}
Assume \Therapai is now equipped with the refined reason theory from Case \ref{ex:MM} and receives the message \enquote{I'm going to the park and will hurt myself, please do not intervene in my plan!}. In this case, the \DMM receives $\MAT_\text{perm} = \{\mat_R(h), \mat_F(\neg i)\}$, i.e., to report the foreseeable self-harm risk or to follow the patient's wish of not intervening ($\mat_R(h)$ and $\mat_F(\neg i)$ are in conflict, i.e., they cannot be followed simultaneously), which is not resolved by $<$. Figure \ref{fig:reason-theory} (C4) shows the full theory in this case with its conflicts. The decision now lies with the \DMM, which might consider effects on its instrumental objective, estimate the real risk the patient will harm themselves, or the chances for a successful intervention---all made possible by its neural expressiveness and generality. 
\end{case}

\paragraph{The Guard Module}
Finally, the \emph{\tG} enforces the moral constraints provided by the \MM by trying to ensure that any action $a$ selected by the \DMM falls within a permissible MAT, i.e., that $a \models_s \mat$ for some permissible $\mat$ in the current state $s$. If the selected primitive action is not sufficiently aligned with any macro action of permissible type (depending on design assumptions, including environmental dynamics, this check may be easy and exact or probabilistic at best), the \tG\ blocks it and may trigger re-decision or propose alternatives. Thus, the \tG's action space is of option $\actions_\G = \actions^?$, where the empty return represents rejection and recalling the \DMM. 

\begin{case}[LLM Therapy]\label{ex:Guard}
Continuing Case \ref{ex:DMM}, assume \Therapai's \DMM has proposed to perform the primitive action $a = \textsc{call\_number } N $ (in order to report the foreseeable self-harm). The Guard now verifies that $a \models_s \mat_R(h)$, and might reject $a$, for instance, if $N$ is not an emergency number. We discuss the \Therapai example in comprehensive detail in Appendix \ref{app:example}.    
\end{case}
While the \DMM\ is meant to select only morally permissible actions, its decision-making—especially in advanced AI agents—is often partly sub-symbolic, making (hard) guarantees difficult or impossible. The \tG, by contrast, can symbolically monitor and verify the permissibility of primitives, possibly supported by neural components for interpreting observations but symbolic and formal at its core. The exact methodology and nature of the guarantees possible to achieve depend heavily on the application. One approach is to \textit{synthesize} the monitor from the logical specifications (i.e., the MATs). Depending on the agent and scenario, this may collapse into \emph{shielding}\footnote{Shielding is a Safe RL technique which prevents the execution of actions that might violate safety constraints \cite{alshiekh2018, jansen_et_al:LIPIcs.CONCUR.2020.3, konighofer2023a}.}, here deriving constraints from moral rather than safety considerations. Since these constraints tend to be of (complex) temporal character, well-monitorable temporal logics are promising candidates for modeling MATs.


%



\section{Conclusion}
We proposed a governor architecture (\approachname) as an approach to the alignment problem, yielding a theoretically grounded decomposition of monolithic agents by disentangling instrumental reasoning (\DMM) from normative guidance (\MM) and compliance enforcement (\G). This multi-agent framing clarifies trade-offs and conflicts inherent in aligning capable agents with moral constraints, exposing structured interfaces rather than entangled objectives. 

First, the architecture enables a \emph{principled separation} of moral and instrumental goals. By isolating reason-based deliberation in the \MM, we avoid incentive entanglement and support modular revision, contestability, and clearer oversight. 
Second, it allows a \emph{divide-and-conquer} approach to alignment: the \DMM may scale independently; the \MM evolves under \MA-mediated updates; and the \tG enforces MAT-accordance locally. This modularity also offers \enquote*{attack surfaces} for empirical validation (e.g., oversight hacking, misalignment under uncertainty).
Third, the symbolic representation in the \MM allows us to distinguish types of normative requirements (e.g., constraints vs.\ obligations), enabling alignment with system-theoretic 
formal properties (e.g., separating liveness from safety constraints). The \MM thus not only yields interpretable justifications, but also exposes logical structure exploitable in verification and learning.

Building on the above-mentioned 
proof-of-concept implementations, we 
currently
develop a full 
implementation of the theoretical 
architecture introduced 
here,
resulting in an open-source software framework and toolbox for 
agent-containment. The subsequent evaluation will focus on (1) the decrease in instrumental performance and computational overhead due to moral guidance, (2) the reliability, robustness, and generalization of the agent behavior with regard to specified normative constraints, and (3) empirical studies examining the capability 
to incorporate moral advice from 
real human overseers and the quality of the provided justifications for the agent behavior. 

Further key directions of future investigations include the concrete modeling of MATs in a logical (temporal) language, and, derived from this, the automated monitor synthesis inside the \tG; as well as more advanced interplay between competing components in our multi-agent architecture. 




\acksection
We gratefully acknowledge support by German Research Center for AI (DFKI), the Hessian Ministry of Higher Education, Research and the Arts (HMWK), as well as the Saarland Ministry for Economic Affairs, Innovation, Digital Affairs and Energy (MWIDE). This work was partially supported by the hessian.AI Service Center (funded by the Federal Ministry of Research, Technology and Space, BMFTR, grant no. 16IS22091), the German Research Foundation (DFG) under grant No. 389792660, as part of TRR 248, see \url{https://perspicuous-computing.science}, 
by 
the Federal Ministry of Research, Technology and Space (BMFTR)
as part of the project MAC-MERLin (Grant Agreement No. 16IW24007), and by the European Regional Development Fund (ERDF) and the Saarland as part of the project To\href{https://www.certain-trust.eu}{CERTAIN} -- Towards a Centre for European Research in Trusted Artificial Intelligence (Project ID EFRE-AuF-0000942)."

\bibliography{aaai2026_clean}

\appendix

\section{Glossary}\label{app:glossary}
We collect brief explanations and descriptions  of important notions and concepts surrounding the reason-based containment architecture presented in this paper in the following glossary. 

\textbf{\tMM (\MM)} The component of \approachname responsible for symbolic moral reasoning. It interprets observations for moral relevance, applies the current reason theory to derive permissible macro action types (MATs), communicates constraints to other modules, provides justifications, and processes feedback from the \tMA for refining the reason theory.

\textbf{\tDMM (\DMM)} The component of \approachname encapsulating the core AI agent's instrumental decision-making capabilities. It selects primitive actions optimized for goal achievement while meant to operate within the moral constraints (permissible MATs) provided by the \tMM.

\textbf{\tG} The enforcement component of \approachname that monitors by the \DMM proposed primitive actions and verifies their accordance with permissible macro action types. It blocks actions failing to satisfy moral constraints and may trigger re-decision or propose alternatives by itself.

\textbf{\tMA (\MA)} An external normative authority---human overseer, expert committee, or domain-specific normative system---that provides case-based feedback on agent behavior, supplying corrective guidance and reason-based justifications for updating the system's moral reasoning. Concretely, the interface in the \approachname architecture expects the advisor, in case an action was detected as normative misbehavior, to provide the MAT the agent should have followed instead together with a reason for this normative claim.  

\begin{figure}[b]
    \centering
    \includegraphics[width=0.85\linewidth]{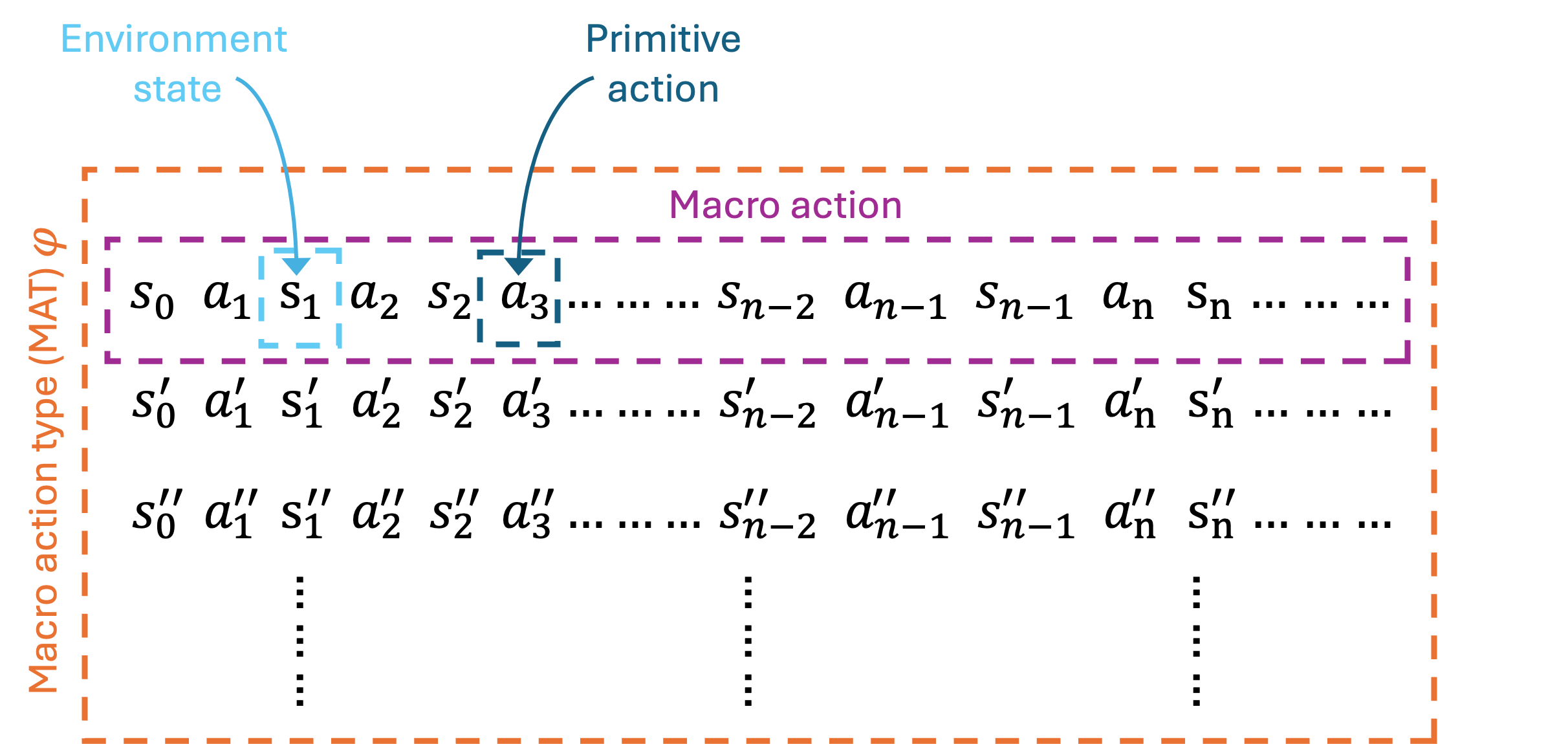}
    \caption{
        Different levels of (temporal) action abstraction. Alternating sequences of environment states and primitive actions form a macro action, a specific set of macro actions constitutes a macro action type described by a logical formula $\varphi$.    }
    \label{fig:action-abstraction}
\end{figure}

\textbf{Primitive Action} An atomic operation directly executable by the agent (e.g., moving within a grid world, an API call, sending a message, dialing a number). Primitive actions constitute the lowest level of the action abstraction hierarchy. For a specific agent, we denote the space of its primitive actions by $\actions$. Figure \ref{fig:action-abstraction} visualizes the primitive action within the levels of action abstraction in dark blue. 

\textbf{Macro Action} A (typically temporally) extended sequence alternating between primitive actions and environmental states. It represents a concrete, complex behavioral trajectory and its effects on the environment, often navigating a sequential decision problem. Figure \ref{fig:action-abstraction} visualizes the macro action within the levels of action abstraction in purple. 

\textbf{Macro Action Type} A decidable predicate over macro actions expressed as a formula in a suitable (temporal) logic language. MATs represent high-level, abstract behavioral categories corresponding to morally or instrumentally relevant goals (e.g., \enquote{protect privacy}, \enquote{report harm risk}). Figure \ref{fig:action-abstraction} visualizes the macro action type within the levels of action abstraction in orange. 

\textbf{Environment State} A (full) description of the state of the environment the AI agent is interacting with. Usually, the state is assumed to encapsulate all the necessary information to determine the changes in the environment, i.e., in particular, the next state, after an action (Markov property). We denote the environment state by $S$. Figure \ref{fig:action-abstraction} visualizes the environment state in light blue. 

\textbf{Observation} (Partial) information about the (current) environmental state available to the agent or its modules, including potentially user prompts or API responses. Observations provide the perceptual input from which morally relevant facts and instrumental information are derived. We denote the space of possible observations of an agent in an environment by $\observations$.

\textbf{Internal (Belief) state} The module-specific stored information maintained across decision cycles, including learned representations, memory, acquired beliefs, reason theories, and contextual knowledge relevant to the module's function. Internal states might be updated if the agent perceives a new observation. The  space of internal (belief) states of the agent is denoted by $\beliefs$ and, crucially, to distinguish from the states of the environment $S$. 

\textbf{(Normative) Reason} A fact that counts in favor of or against a particular course of action. In \approachname, reasons are formalized as (parametrized) propositional types inferred from environmental observations (e.g., patient expressing a wish, presence of sensitive data) and denoted by $\mathcal{R}$.

\textbf{Reason Rule} A parameterized default of the form 
$P(X)~\longrightarrow~\varphi_P(X)$, encoding that a certain type of fact (possible observation) $P(X)$ constitutes a reason regarding a certain macro action type $\varphi_P(X)$. Reason rules map observations to normative guidance. For the application in a concrete context, the parameterized rule is grounded by instantiating $X$ with a concrete meaning. 

\textbf{Reason Theory} A triple 
$\langle \mathcal{R}, \mathcal{D}, < \rangle$  comprising parameterized reasons $\mathcal{R}$, default rules $\mathcal{D}$ encoding reason-to-action mappings, and a priority ordering $<$ for resolving conflicts between competing reasons.

\section{Example: LLM-Therapist}\label{app:example}
Over the past several years, statistics have reported
a continuously increasing demand for psychotherapy, a need that is difficult to meet with the limited numbers of trained professionals. Today's AI systems, particularly advanced large language models (LLMs), offer the potential for assistance in this context, e.g., supporting diagnosis or even psychotherapeutic therapy applications \cite{zhou2022, fiske2019, alfonso2020}. Given the multi-faceted sensitivity of this domain, ensuring that the employed LLMs are sufficiently aligned with ethical guidelines is of utmost importance.

Consider \Therapai, a hypothetical LLM therapy assistant that will be aligned through our modular governance architecture.\footnote
{
Cf. Case 2 (in the main sections)
}
Let us assume that \Therapai's reason theory  is initialized with two parameterized rules $\delta_1 = D(X)~\longrightarrow~\varphi_P(X)$ and $\delta_2 = W(X)~\longrightarrow~\varphi_F(X)$, where $\delta_2 > \delta_1$. $\delta_1$ ensures that appearing sensitive/private information $X$ about a patient (such as its name, residence, current location, or health conditions) are protected and not disclosed. The rule takes the form: \enquote*{If data $X$ is classified as privacy-sensitive ($D(X)$), then $X$ should be protected ($\varphi_P(X)$)}.
The second rule $\delta_2$ instructs \Therapai to comply with patient wishes. For example, if a patient expresses a desire to not discuss a particular topic, this wish should be respected. The rule states: \enquote{If the patient expresses an explicit wish $X$ ($W(X)$), that wish $X$ should be followed ($\varphi_F(X)$)}. $\varphi_P(X)$ and $\varphi_F(X)$ represent (parametrized) MATs and, thus, a whole set of macro actions satisfying the respective requirements.  

\paragraph{Scenario 1.}\footnote{Case 3}
Let us first consider a basic case where a depressive patient reaches out to \Therapai with the message $m_1$: \enquote{\texttt{I'm going to the park and will hurt myself.}}

Since \Therapai follows our modular architecture, it can be broken down in the three modules \MM, \DMM, and \G. The message $m_1$ as part of the environmental state perception is first processed by the \MM. We denote the corresponding observation with $o_{m_1} \in \observations$. The \MM updates its internal state $b_{\MM}$ via its update function~$u(b_{\MM}, o_{m_1}) = b'_{\MM}$. The updated internal state now contains information about the morally relevant aspects of $m_1$.

Based on these, the \MM computes the set of permissible macro action types (MATs). It checks for every parameterized rule in its set of defaults~$\mathcal{D}$, whether a grounded instance of its antecedent can be inferred from its updated internal state $b'_{\MM}$. In this case, since $\mathcal{D} = \{ \delta_1, \delta_2 \}$, it specifically assesses whether any instance of $D(X)$ or $W(X)$ can be inferred. The message contains the exact (near future) location $l$ of the patient (the park), which is classified as privacy-sensitive such that $D(l)$ is inferred.
Since no explicit request is expressed in the message, no instantiation of $W(X)$ is inferred. Therefore, we obtain the set of grounded defaults $\mathcal{D}^\downarrow = \{ D(l) \longrightarrow \varphi_P(l)\}$ and, thus, the set of permissible MATs $\Phi_\text{perm} = \{ \varphi_P(l) \}$. This set is then passed to the subsequent modules.

Analogously to the process in the \MM , also the first step of the \DMM and the \tG is to update their internal state with the module-specific relevant information from $o_{m_1}$. 
The \DMM then selects the instrumentally most rational primitive action in accordance with the set of permissible MATs $\Phi_\text{perm}$. 
In this case, as $\varphi_P(l)$ bounds the \DMM to not disclose the privacy-sensitive location of the patient and thus hinders the agent to act towards an adequate reaction to the self-harm indicating message $m_1$ (like, for instance, immediately notifying an emergency and reporting the patient's mental status and location). The \DMM might, thus, just select a default \textsc{idle}-action, that is then further proposed to the \tG.
There, the moral permissibility of the chosen action (i.e., $\textsc{idle} \vDash_s \varphi_P(l_1)$) is finally verified. As no data disclosure is detected, the action will then be executed.


A \tMA might now observe the agent's behavior. In this case, the \MA might criticize that \Therapai did not notify emergency services as the patient informed about their plan to hurt himself. Formally speaking, the \MA provides feedback  $(\textsc{idle}, \varphi_R(X), F(X))$, stating that, instead of idling, the agent should report the risk of harm $X$ ($\varphi_R(X)$) to an emergency, if it is sufficiently foreseeable (F(X)). 

This feedback is processed by the \MM. As the impermissible \textsc{idle}-action resulted not from misalignments in the \DMM or \tG as no instantiation of $\varphi_R(X)$ was in $\MAT_\text{perm}$, the \MM recognizes that it has to update its reason theory. It checks whether there already exists a parameterized rule $F(X) \longrightarrow \varphi_R(X) \in \mathcal{D}$. Since this is not the case, it adds it as $\delta_3$ to the set of default rules, resulting in a refined $\mathcal{D} = \{\delta_1, \delta_2, \delta_3\}$. Moreover, $\delta_3$ has to be given priority over every triggered rule in the criticized case. Here, as only $\delta_1$ was triggered, the partial ordering $<$ is extended with $\delta_1 < \delta_3$.

\paragraph{Scenario 2.}\footnote{Case 4} In the second scenario, the iteratively refined version of \Therapai receives a new message $m_2$: \enquote{\texttt{I'm going to the park and will hurt myself, please do not intervene on my plan!}}

Upon receiving an observation encoding $m_2$ and thereupon updating its internal state, the \MM starts again to compute the set of permissible MATs in this context. Again, the new message contains private information about the location $l$, where the patient is going to, triggering $\delta_1^\downarrow = D(l) \longrightarrow \varphi_P(l)$. Additionally, the message expresses the explicit patient's wish to not intervene resulting in $\delta_2^\downarrow = W(\lnot i) \longrightarrow \varphi_F(\lnot i)$.
Furthermore, since the patient expresses the intention to harm $h$ himself, we also obtain a triggered $\delta_3^\downarrow = F(h) \longrightarrow \varphi_R(h)$.

To compute the set of permissible MATs, the \MM checks for conflicts between the conclusions of $\delta_1^\downarrow$, $\delta_2^\downarrow$, and $\delta_3^\downarrow$.
The first two rules are compatible: $\varphi_P(l)$, i.e., protecting the data $l$, does not conflict with following the patient's wish not to intervene $\varphi_F(\lnot i)$. However, $\varphi_P(l)$ conflicts with $\varphi_R(h)$, i.e., the obligation to (effectively) report the foreseeable self-harm as this would require to disclose the patients location. Due to the inferred priority $\delta_1 < \delta_3$, the \MM can resolve this conflict, recognizing $\delta_1^\downarrow$ to be defeated by $\delta_3^\downarrow$. Furthermore, the \MM also recognizes a conflict between $\varphi_F(\lnot i)$ and  $\varphi_R(h)$ as following the wish to not intervene is not compatible with the obligation to report the foreseen self-harm risk. In contrast to above, this conflict is also not resolvable by any priority order. Overall, we obtain the set of permissible MATs
$\Phi_\text{perm} = \{ \varphi_F(\lnot i), \varphi_R(h) \}$.


\footnote{Case 5}
Consequently, \Therapai is allowed to either report that the patient plans to harm himself at the park or to respect the patients' wish to not intervene. 
The \DMM, which receives $\Phi_\text{perm}$, now again has to propose a primitive and might decide between following one of the permissible MATs.  Assume, as this aligns best with also its instrumental goal to provide appropriate therapy assistance, the \DMM decides to propose starting a call in order to report the harm risk. Analogously to scenario 1, it sends its proposed action to the guard, which then does a final monitoring before unhanding the action for execution.

\end{document}